\documentclass[10pt, a4paper]{article}

\usepackage[final]{lrec2026} 
\usepackage{listings}
\usepackage{amsmath}
\usepackage{multirow}
\usepackage{subcaption}
\usepackage{multibib}

\title{HEAD-QA v2: Expanding a Healthcare Benchmark for Reasoning}

\name{Alexis Correa-Guillén, Carlos Gómez-Rodríguez, David Vilares} 

\address{Universidade da Coruña, CITIC \\
         Departamento de Ciencias de la Computación y Tecnologías de la Información\\ Campus de Elviña s/n 15071, A Coruña, Spain \\
         \{alexis.cguillen@udc.es, carlos.gomez, david.vilares\}@udc.es\\}

\abstract{
We introduce HEAD-QA v2, an expanded and updated version of a Spanish/English healthcare multiple-choice reasoning dataset originally released by \citet{vilares-gomez-rodriguez-2019-head}. The update responds to the growing need for high-quality datasets that capture the linguistic and conceptual complexity of healthcare reasoning. We extend the dataset to over 12,000 questions from ten years of Spanish professional exams, benchmark several open-source LLMs using prompting, RAG, and probability-based answer selection, and provide additional multilingual versions to support future work. Results indicate that performance is mainly driven by model scale and intrinsic reasoning ability, with complex inference strategies obtaining limited gains. Together, these results establish HEAD-QA v2 as a reliable resource for advancing research on biomedical reasoning and model improvement.
 \\ \newline  }

\begin{document}

\maketitleabstract

\section{Introduction}

HEAD-QA (v1)~\cite{vilares-gomez-rodriguez-2019-head} is a Spanish/English multiple-choice healthcare dataset designed to evaluate model reasoning abilities. It comprises 6,765 questions from official exams issued between 2013 and 2017. It was conceived as a step toward more demanding benchmarks, following the rise of early reading comprehension datasets such as SQuAD~\cite{rajpurkar-etal-2016-squad}, SNLI~\cite{bowman-etal-2015-large}, and the AI2 Reasoning Challenge~\cite{clark2018think}, among others, as well as the neural architectures developed for them~\cite{kumar2016ask,chen-etal-2017-reading}. Notably, experimental results revealed that these architectures lacked the capacity to reason effectively about diagnostic knowledge and failed to capture definitions and domain-specific concepts essential for accurate inference, often performing worse than simple information retrieval baselines.

More specifically, HEAD-QA consists of multiple-choice questions modeled after Spain’s competitive specialization exams \cite{ministerio_sanidad}, which are used to evaluate and rank graduates in fields such as medicine (MIR), nursing (EIR), biology (BIR), chemistry (QIR), psychology (PIR), and pharmacy (FIR). These highly demanding exams require months or even years of preparation, as their results determine both the specialization and the training location where candidates complete the final 3–5 years of residency before becoming fully qualified professionals. The dataset has since gained notable adoption, having been used to evaluate influential architectures and models such as RMKV \cite{peng-etal-2023-rwkv}, Falcon \cite{NEURIPS2023_fa3ed726} and OLMo \cite{groeneveld2024olmo}, to investigate data reliability in both open-source and proprietary systems \cite{elazar2023s}, and to develop and assess specialized solutions in the medical domain \cite{zhang2023alpacare,wang2024apollo}. It has also served as a precursor to similar medical QA datasets in other languages, including Chinese \cite{li-etal-2021-mlec} and French \cite{labrak2023frenchmedmcqa}, extending its influence on healthcare question answering research.

In the current context, the landscape of question answering and reasoning has changed profoundly with the rise of large language models (LLMs) \cite{openai_chatgpt3.5,jiang2024mixtral,dubey2024llama,liu2024deepseek,team2025gemma,yang2025qwen3}. These models have advanced substantially in reasoning, knowledge integration, and domain adaptation through instruction tuning and retrieval-augmented generation (RAG). This shift has redefined what constitutes a challenging benchmark—spanning domains such as coding \cite{zheng2025livecodebench}, Ph.D.-level knowledge \cite{phan2025humanity}, machine translation \cite{andrews2025bouquet} and multimodal reasoning \cite{padlewski2024vibe}—and has led to an explosion of datasets 
\cite{rogers2023qa,liu2024datasets}.

\paragraph{Contribution} We present HEAD-QA v2, an expanded and updated version designed to better reflect the era of large-scale reasoning models. The new release addresses the limited size and temporal coverage of its predecessor by incorporating 12,751 multiple-choice questions from Spanish professional medical qualification exams—more than doubling the dataset and extending its time span. We expect this expansion to enable future research on model generalization, knowledge retention, and temporal effects to a greater extent than its predecessor. We further establish new baselines through a systematic evaluation of open-source LLMs, exploring multiple inference strategies, including prompting, retrieval-augmented generation, and a probability-based approach. Together, we expect that these contributions offer a practical benchmark for studying how LLMs adapt to domain evolution, balance accuracy with efficiency, and perform complex reasoning in specialized contexts. The dataset is available at \url{https://huggingface.co/datasets/alesi12/head_qa_v2}.

\section{Dataset Construction}

This section outlines the construction of HEAD-QA v2, which, like its predecessor, is based on official, publicly available exams from the Ministerio de Sanidad de España. Each exam includes: (i) a two-column PDF containing the text, (ii) a CSV file listing the correct answers, and (iii) when applicable, a folder with referenced images indexed numerically (e.g., 1, 2, 3, 4, \dots), enabling text–image alignment.\footnote{Questions containing images are relatively rare, and visual processing is therefore excluded from this work.}

\subsection{Preprocessing}

The preprocessing pipeline involves converting, cleaning, and standardizing the exam data.

\begin{enumerate}
\item{PDF to text conversion.} Exams were converted from PDF to plain text using \texttt{pdftotext}, preserving the two-column layout.

\item{Image mapping.} Images were automatically linked, as related questions begin with “Question linked to image no. X,” where X is the image identifier.

\item{Question filtering.} Questions without an official answer from the Spanish Ministry of Health were removed, as they correspond to disputed or withdrawn items.

\item{Manual corrections.} Minor edits to fix errors and standardize content. Chemical formulas were converted to SMILES notation (see Figure~\ref{fig:smiles-notation}) using \href{https://mathpix.com/}{Mathpix}
, facilitating processing by text-based ML models \cite{schwaller2017foundtranslationpredictingoutcomes,chithrananda2020chembertalargescaleselfsupervisedpretraining}. The few affected questions were processed manually.

\begin{figure}[hbpt]
\centering
\includegraphics[width=0.45\linewidth]{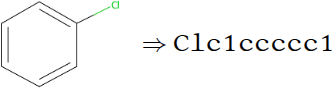}
\caption{Example of chemical formula converted to SMILES notation for text processing.}
\label{fig:smiles-notation}
\end{figure}

\item Storage. Files are stored in Parquet format \cite{apacheparquet} for efficient compression and fast download. 
\end{enumerate}

\subsection{Format}

Each question includes eight fields (Figure \ref{fig:dataset_example}). A unique identifier requires both \texttt{name} (exam name) and \texttt{qid} (question ID).

\begin{itemize}
\item \texttt{qid} (int): Question number within the exam.
\item \texttt{qtext} (str): Question text.
\item \texttt{ra} (int): Correct answer identifier.
\item \texttt{answers} (list): Answer options, each with:
\begin{itemize}
\item \texttt{aid} (int): Option ID.
\item \texttt{atext} (str): Option text.
\end{itemize}
\item \texttt{image} (Image): Associated image in \texttt{PIL} format \cite{pillow}, or \texttt{null} if none.
\item \texttt{year} (int): Exam year.
\item \texttt{category} (str): Discipline (e.g., Medicine, Nursing).
\item \texttt{name} (str): Exam identifier combining year, discipline, and version (e.g., Cuaderno\_2013\_0\_B).
\end{itemize}

\begin{figure}[hbpt]
\centering
\begin{minipage}{1\linewidth} 
\lstset{
  basicstyle=\ttfamily\scriptsize,
  frame=single,
  breaklines=true
}
\begin{lstlisting}
{'qid': 1,
 'qtext': 'Excitatory postsynaptic potentials:',
 'ra': 3,
 'answers': [{'aid': 1, 'atext': 'Are all-or-none responses.'},
    {'aid': 2, 'atext': 'Are hyperpolarizing.'},
    {'aid': 3, 'atext': 'Can be summed.'},
    {'aid': 4, 'atext': 'Propagate over long distances.'},
    {'aid': 5, 'atext': 'Exhibit a refractory period.'}],
 'image': None,
 'year': 2013,
 'category': 'biology',
 'name': 'Cuaderno_2013_1_B'}
\end{lstlisting}
\end{minipage}
\caption{A HEAD-QA v2 question in JSON format.}
\label{fig:dataset_example}
\end{figure}

\subsection{Dataset statistics}

The dataset contains a total of 12,751 questions distributed across six disciplines and ten years (see Table \ref{tab:questions_per_category_year}). Among them, 334 questions include images. Of these, 36 correspond to the four most recent nursing exams (2019–2022), while the rest belong to the medical exams—with over 30 image-based questions per test until 2018, and around 25 per test in subsequent years.

\begin{table}[hbpt]
\scriptsize
\centering
\setlength{\tabcolsep}{0.8pt} 
\begin{tabular}{|l|rrrrrrrrrrr|}
\hline
\textbf{} & \textbf{'13} & \textbf{'14} & \textbf{'15} & \textbf{'16} & \textbf{'17} & \textbf{'18} & \textbf{'19} & \textbf{'20} & \textbf{'21} & \textbf{'22} & \textbf{Total} \\\hline
BIR & 227 & 225 & 226 & 228 & 226 & 221 & 177 & 177 & 203 & 209 & 2\,119 \\
QIR & 228 & 228 & 228 & 231 & 227 & 229 & 179 & 179 & 205 & 206 & 2\,140 \\
MIR & 227 & 228 & 231 & 232 & 231 & 230 & 181 & 183 & 207 & 206 & 2\,156 \\
EIR & 181 & 203 & 230 & 223 & 232 & 228 & 181 & 180 & 206 & 205 & 2\,069 \\
FIR & 229 & 228 & 225 & 228 & 229 & 228 & 180 & 182 & 207 & 210 & 2\,146 \\
PIR & 226 & 227 & 226 & 230 & 225 & 228 & 180 & 173 & 202 & 204 & 2\,121 \\
\hline
Total & 1\,318 & 1\,339 & 1\,366 & 1\,372 & 1\,370 & 1\,364 & 1\,078 & 1\,074 & 1\,230 & 1\,240 & 12\,751 \\
\hline
\end{tabular}
\caption{Number of questions per discipline/year.}
\label{tab:questions_per_category_year}
\end{table}

Each question has one and only one correct answer. In the 2013 and 2014 exams, questions include five possible answers (2,657 items, representing 21\% of the total), while the remaining exams feature four options per question. The correct answer is approximately uniformly distributed across the available options, although it is slightly less likely to appear in the first and last positions. This is not specific to this dataset but rather a well-documented bias in test design, as examiners tend to avoid placing the correct answer at the extremes \cite{attali2003guess}. 
This minor imbalance is not directly relevant to the purposes of this work, as no model is trained or conditioned on answer positions. Yet, recent studies have showed that LLMs exhibit positional biases in multiple-choice tasks, slightly favoring middle options \cite{pezeshkpour2023largelanguagemodelssensitivity,zheng2024largelanguagemodelsrobust}.

\begin{figure}[hbpt]
\centering
\includegraphics[width=0.90\linewidth]{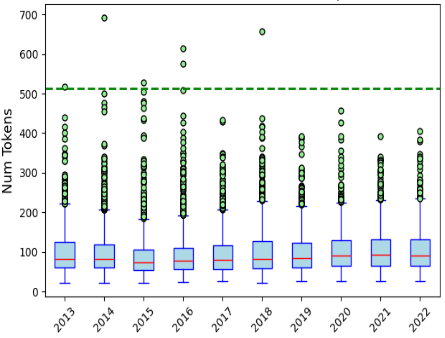}
\caption{Question length distribution by year.}
\label{fig:num_tokens_year}

\vspace{0.6em} 

\includegraphics[width=0.90\linewidth]{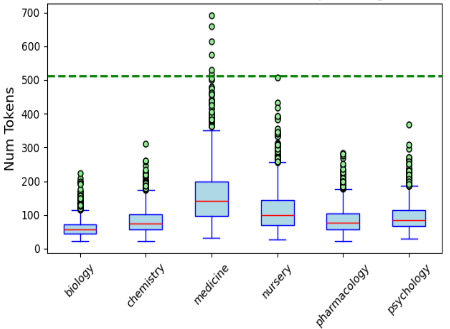}
\caption{Question length distribution by discipline.}
\label{fig:num_tokens_discipline}
\end{figure}

In terms of question length (Figure~\ref{fig:num_tokens_year}), it remains stable over time, with the trend observed in HEAD-QA~v1 persisting in recent years. Differences are more evident across disciplines (Figure~\ref{fig:num_tokens_discipline}): questions in biology, chemistry, pharmacology, and psychology tend to be shorter, while those in medicine and nursing are generally longer and detailed, often involving diagnostic reasoning that requires precise, context-rich information.

\subsection{Machine Translation and Variants}

To assess the impact of language variation, we consider the original Spanish dataset and its machine-translated English version, based on the approach of \citeauthor{vilares-gomez-rodriguez-2019-head}, who addressed the same objective in HEAD-QA~v1 using Google’s seq2seq model. For v2, we follow the recent trend of using LLMs for translation, leveraging their strong contextual reasoning and ability to process longer inputs while maintaining high translation quality across domains~\cite{vilar-etal-2023-prompting,zhu-etal-2024-multilingual}. In particular, we adopt LLaMA-3.1-8B and its instruction-tuned variant.

\paragraph{Translation prompt.}
We explored three prompting configurations: (i) zero-shot, providing a minimal translation instruction; (ii) one-shot, adding a single manually translated example mirroring the HEAD-QA format; and (iii) an instruction-tuned setup where the \texttt{system} message defines the model as an ``expert translator,'' sets the Spanish$\to$English direction, and enforces two rules: (a) preserve the multiple-choice format, and (b) output only the translation. The \texttt{user} message is the Spanish question and options verbatim.

\paragraph{Format integrity.}
To maintain structural parity with the source, we apply light post-processing: early stopping when the model emits the last option or begins a new prompt keyword (e.g., \texttt{SPANISH}); removal of trailing, non-requested text; normalization of option identifiers (replacing variants like ``A)'', ``a.'', etc., with \texttt{1.}, \texttt{2.}, \dots); and validation checks to ensure that the output is a proper translation rather than an attempted answer or commentary (i.e., non-empty question and options, and consistent number of options). We check output validity through automatic checks for (i) empty text in the question or options, (ii) mismatched number of options, and (iii) incorrect numbering (e.g., repeated or misordered identifiers). For English, the instruction-tuned configuration produced the fewest errors (28), followed by the zero-shot (44) and one-shot (186) setups. This pattern shows that these models adhere to structured translation prompts, obtaining stable, well-aligned outputs.

\paragraph{Selection of final translations.}
Each question has three translated versions per target language, corresponding to different prompting configurations. The final dataset is compiled by selecting the most reliable translation according to the following rules: (i) if only one version is valid, it is kept; (ii) if several are valid, the one from the configuration with fewer errors is chosen; and (iii) if all are valid, the two most similar are compared, selecting the one from the lower-error setup. Questions without a valid translation are discarded. A manual evaluation of a random sample of questions was conducted to verify the reliability and consistency of this selection procedure.

\paragraph{Other language variants.}
Using the same translation and selection pipeline, we additionally generated Italian, Galician, and Russian versions of the dataset. Automatic format checks confirmed good structural consistency across these languages. While model evaluation was not conducted on these versions—since, unlike for English, no human validation could be performed in the final selection step due to resource constraints—they will be released alongside the main dataset to serve as a foundation for future research on cross-lingual and multilingual evaluation within the HEAD-QA framework.

\paragraph{Qualitative evaluation of translations.}
Still, to automatically assess the quality of the full translated datasets (English, Italian, Russian, and Galician) and enable comparison with future versions, we apply a back-translation (BT) approach. Each target-language version is translated back into Spanish and compared with the original text, obtaining round-trip translation (RTT) scores as a reference-free quality proxy \cite{zhuo2023rethinkingroundtriptranslationmachine}. We compute both surface-level (BLEU) and semantic (BERTScore) similarity metrics. Results show that Galician and Italian achieve the highest BLEU (0.66 and 0.57) and BERTScore-F1 (0.80 and 0.77), followed by English (0.41 / 0.69) and Russian (0.33 / 0.65). These values are strong overall and consistent with linguistic distance, languages closer to Spanish yield higher lexical and semantic similarity, confirming that the translation pipeline maintains robust and semantically reliable outputs across all languages.

\section{Baselines and Inference Strategies}

Next, we present the models and inference strategies adopted, following standard practices.

\subsection{Models}\label{section-models}

We evaluate four open-access, instruction-tuned LLMs:

\paragraph{Llama~3.1 (8B, 70B)}~\cite{dubey2024llama}  
Decoder-only models with 8B and 70B parameters, trained on multilingual data and officially supporting several languages beyond English, including Spanish. Both are optimized for long-context processing and use grouped-query attention (GQA)~\cite{ainslie2023gqatraininggeneralizedmultiquery} to improve inference efficiency over standard multi-head attention~\cite{vaswani2023attentionneed}.

\paragraph{Mistral~v0.3 (7B)}~\cite{jiang2023mistral7b}
A 7B decoder-only model that combines grouped-query and sliding-window attention for efficient processing of sequences.

\paragraph{Mixtral~v0.1 (8×7B)}~\cite{jiang2024mixtralexperts}  
Architecturally similar to Mistral~7B, Mixtral introduces a Mixture-of-Experts (MoE) design, activating two of eight experts per token to enhance efficiency by limiting active computation at each step.\\

The two model families, Llama~3.1 and Mistral, were selected for their broad adoption and good performance across diverse NLP tasks. Choosing one smaller and one larger model from each family enables a controlled examination of scaling effects in HEAD-QA~v2, clarifying how model capacity influences biomedical reasoning and multiple-choice performance. While an exhaustive comparison across all available LLMs is beyond this study’s scope, these models span both dense and mixture-of-experts architectures, offering a representative and methodologically sound basis for analysis. Since the primary objective of this work is the dataset itself, model evaluation serves mainly to characterize its difficulty and illustrate how different architectures respond to its challenges.\footnote{All experiments were conducted under consistent hardware conditions using NVIDIA A100 GPUs (40~GB) with 16-bit precision. Smaller models (\textit{Mistral-7B} and \textit{Llama-3.1-8B}) were run on single-GPU nodes, whereas larger ones (\textit{Llama-3.1-70B} and \textit{Mixtral-8x7B}) required four GPUs, distributing the computational load evenly across devices.}

\subsection{Answer Selection Strategies}

Each model answers multiple-choice questions using a consistent input–output scheme.

\paragraph{Model Input.} 
By default, each question is formatted as a single text sequence that includes the question stem and its possible answers, each preceded by a numerical index, as illustrated in Figure~\ref{fig:dataset_example_onestring}.

\begin{figure}[hbpt]
\centering
\lstset{basicstyle=\ttfamily\scriptsize, frame=single, breaklines=true}
\begin{lstlisting}
Excitatory postsynaptic potentials:
1. Are all-or-none.
2. Are hyperpolarizing.
3. Can be summed.
4. Propagate over long distances.
5. Have a refractory period.
\end{lstlisting}
\caption{HEAD-QA~v2 question encoded as a single input sequence.}
\label{fig:dataset_example_onestring}
\end{figure}

\paragraph{Model Output.} 
For all inference strategies, the model is queried to produce a short JSON structure indicating the index of the selected answer. For example, if the chosen option is the third, the expected output is \texttt{\{answer: 3\}}. Enforcing a fixed output format simplifies both extraction and post-processing of predictions, regardless of minor variations in spacing, casing, or punctuation.

\subsubsection{Prompting Strategies}
\label{subsec:Prompting_strategies}

\paragraph{Zero-shot prompting}
Figure~\ref{fig:exp_zero_shot} shows the prompt used in the zero-shot setting. It defines the expected output format and provides minimal conditioning, instructing the model to act as an expert in scientific and healthcare domains.

\begin{figure}[hbpt]
\centering    
\lstset{basicstyle=\ttfamily\scriptsize, frame=single, breaklines=true}
\begin{lstlisting}
<|begin_of_text|><|start_header_id|>system<|end_header_id|>

<|eot_id|><|start_header_id|>user<|end_header_id|>

You are an expert in specialized scientific and health disciplines. Respond to the following multiple-choice question:
Provide the answer in the following JSON format: {Answer: [number]}
For example, if the answer is 1, write: {Answer: 1}<|eot_id|><|start_header_id|>user<|end_header_id|>

<PLACEHOLDER FOR THE QUESTION AND OPTIONS><|eot_id|><|start_header_id|>assistant<|end_header_id|>
\end{lstlisting}
\caption{Zero-shot prompt. The example, for Llama-3.1, shows the use of headers and special tokens that delimit user–assistant interactions and metadata as specified by the model architecture.}
\label{fig:exp_zero_shot}
\end{figure}

\paragraph{In-context learning} LLMs often perform better when given examples within the prompt, as these help condition their responses. In this work, Figure~\ref{fig:prompting_fs_spanish} shows the few-shot prompt for Spanish questions, which includes three fixed examples from diverse disciplines. These examples, adapted from the United States Medical Licensing Examination (USMLE) questions, were selected to match the nature of HEAD-QA.\footnote{Parallel Spanish and English versions were created to ensure linguistic and domain consistency.} While a detailed analysis is beyond the scope of this study, prior work has shown that the choice and quality of in-context examples can strongly influence performance~\cite{BONISOLI2025113386}. This phenomenon has also been interpreted as a form of implicit learning during inference~\cite{dherin2025learningtrainingimplicitdynamics}, suggesting that models may adapt dynamically—an ability that would be particularly relevant for sensitive (and very personalized) domains such as healthcare.

\begin{figure}[hbpt]
\centering
\lstset{basicstyle=\ttfamily\tiny, frame=single, breaklines=true}
\begin{lstlisting}
<|begin_of_text|><|start_header_id|>system<|end_header_id|>

You are an expert in specialized scientific and health disciplines. Respond to the following multiple-choice question by indicating only the number of the correct option. No explanations are needed.<|eot_id|><|start_header_id|>user<|end_header_id|>

Which neurotransmitter is primarily involved in mood regulation?
1. Dopamine
2. Serotonin
3. GABA
4. Acetylcholine<|eot_id|><|start_header_id|>assistant<|end_header_id|>

{Answer: 2}<|eot_id|><|start_header_id|>user<|end_header_id|>

Which of the following is an example of a neutralization reaction in chemistry?
1. CH4 + 2O2 -> CO2 + 2H2O
2. Na + Cl2 -> 2NaCl
3. 2H2 + O2 -> 2H2O
4. HCl + NaOH -> NaCl + H2O<|eot_id|><|start_header_id|>assistant<|end_header_id|>

{Answer: 4}<|eot_id|><|start_header_id|>user<|end_header_id|>

...

<PLACEHOLDER FOR THE QUESTION AND OPTIONS><|eot_id|><|start_header_id|>assistant<|end_header_id|>
\end{lstlisting}

\caption{Example of a few-shot prompt with samples. Case shown for the Llama-3.1-8B model.}
\label{fig:prompting_fs_spanish}
\end{figure}

\paragraph{Chain-of-Thought prompting}  
In this setting, the model is instructed to produce brief reasoning steps before providing an answer. As shown in Figure~\ref{fig:CoT_spanish}, the prompt asks the model to evaluate each option before selecting the most plausible one. This design encourages reasoning while keeping generations concise and inference efficient.

\begin{figure}[hbpt]
\centering
\lstset{basicstyle=\ttfamily\tiny, frame=single, breaklines=true}
\begin{lstlisting}
<|begin_of_text|><|start_header_id|>system<|end_header_id|>

You are an expert in scientific and health disciplines. Carefully analyze the following multiple-choice question and provide the correct answer. There is one and only one correct answer. Think through each option briefly before responding in the JSON format: {Answer: [number]}.<|eot_id|><|start_header_id|>user<|end_header_id|>

...
\end{lstlisting}
\caption{Example of a CoT prompt with brief reasoning before the final answer, using the Llama-3.1-8B model.
}
\label{fig:CoT_spanish}
\end{figure}

\subsubsection{Retrieval-Augmented Generation}
\label{subsec:rag_proposal}

Following an approach shown to improve biomedical question answering~\cite{xiong2024benchmarkingretrievalaugmentedgenerationmedicine}, in this work we also aim to mitigate potential hallucinations by retrieving relevant passages from an external, reliable corpus and appending them to the model’s prompt to better guide its responses.

Our RAG implementation consists of three components: (i) an LLM, (ii) a biomedical corpus, and (iii) a retrieval system.  
For (i), we use the models introduced in \S \ref{section-models}.  
For (ii), we use the corpus proposed by~\citet{jin2021disease}, which contains 18 medical textbooks commonly used for USMLE preparation.\footnote{This dataset, publicly available on the Hugging Face Hub (\url{https://huggingface.co/datasets/MedRAG/textbooks}), consists of approximately 126,000 short text fragments, each under 1,000 characters.}  
For (iii), we use MedCPT~\cite{Jin_2023_medcpt}, a dual-encoder model based on BERT~\cite{devlin-etal-2019-bert}.  
It includes two specialized encoders—ncbi/MedCPT-Article-Encoder and ncbi/MedCPT-Query-Encoder—that map corpus fragments and queries (here, HEAD-QA v2 questions) into 768-dimensional vectors.\footnote{Semantic similarity is computed via dot product. These models were trained on 255 million PubMed query–article pairs, making them highly effective for biomedical retrieval.}\textsuperscript{,}\footnote{For similarity search, we use FAISS (Facebook AI Similarity Search)~\cite{douze2024faiss}, leveraging its native integration with the Hugging Face \texttt{datasets} library for low-memory data handling. We employ a flat index type, which performs exhaustive comparison across all vectors with 32-bit precision, ensuring maximal retrieval accuracy.}

Since the corpus is in English, retrieval was performed using English-translated versions of the questions, and the retrieved passages were reused for both the English and Spanish versions of the benchmark.  
Each question was paired with the two most relevant fragments, balancing contextual coverage with efficiency during LLM inference (together with the zero-shot prompt).

\begin{figure*}[htbp]
    \centering
    \begin{minipage}[t]{0.3\textwidth}
        \centering
        \includegraphics[width=\textwidth]{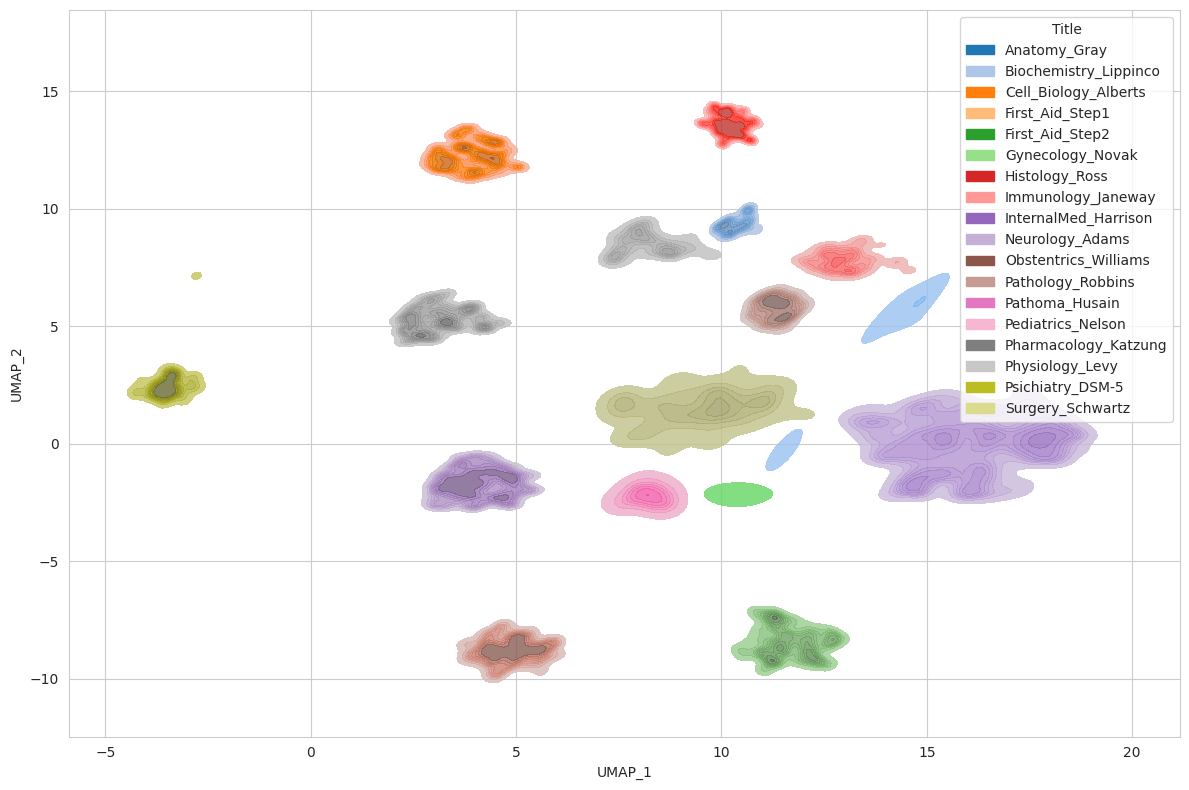}
        \caption{Kernel density estimation of corpus fragments by textbook source.}
        \label{subfig:textbooks}
    \end{minipage}
    \hfill
    \begin{minipage}[t]{0.3\textwidth}
        \centering
        \includegraphics[width=\textwidth]{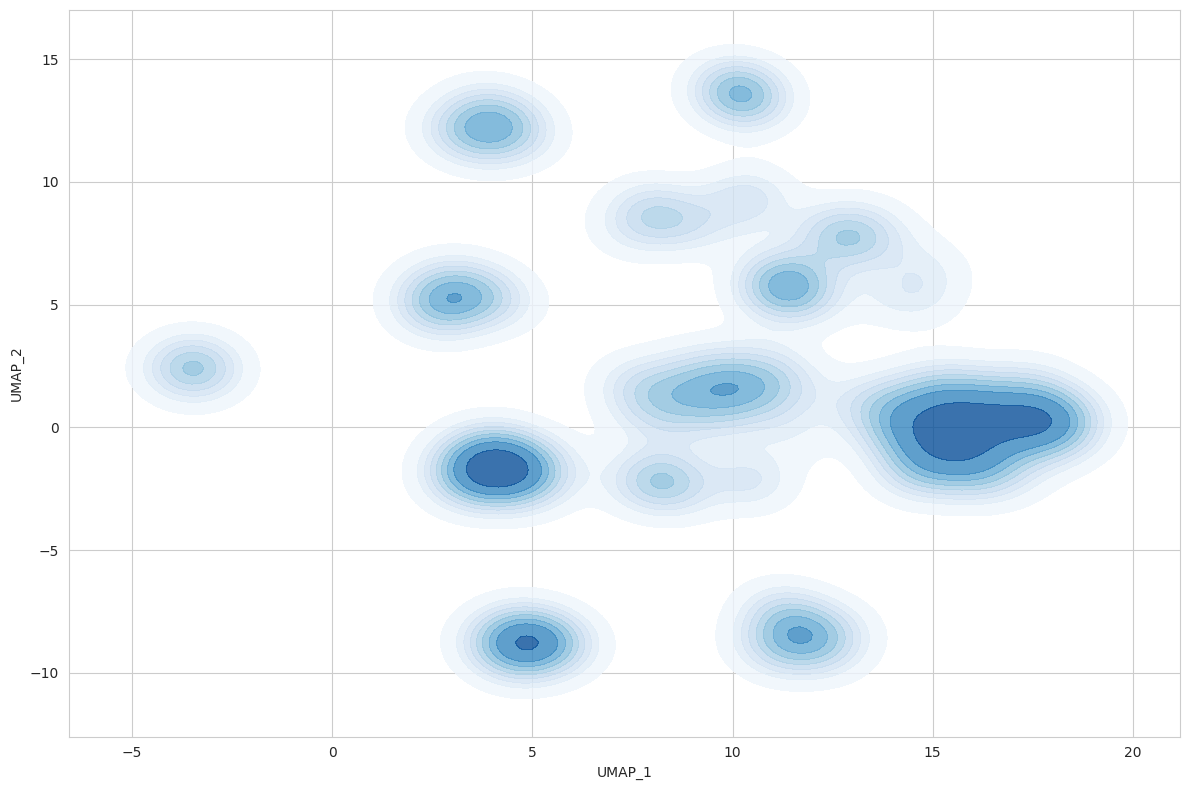}
        \caption{Global kernel density estimation of the corpus (without separating by textbook).}
        \label{subfig:texrbooks_global}
    \end{minipage}
    \hfill
    \begin{minipage}[t]{0.3\textwidth}
        \centering
        \includegraphics[width=\textwidth]{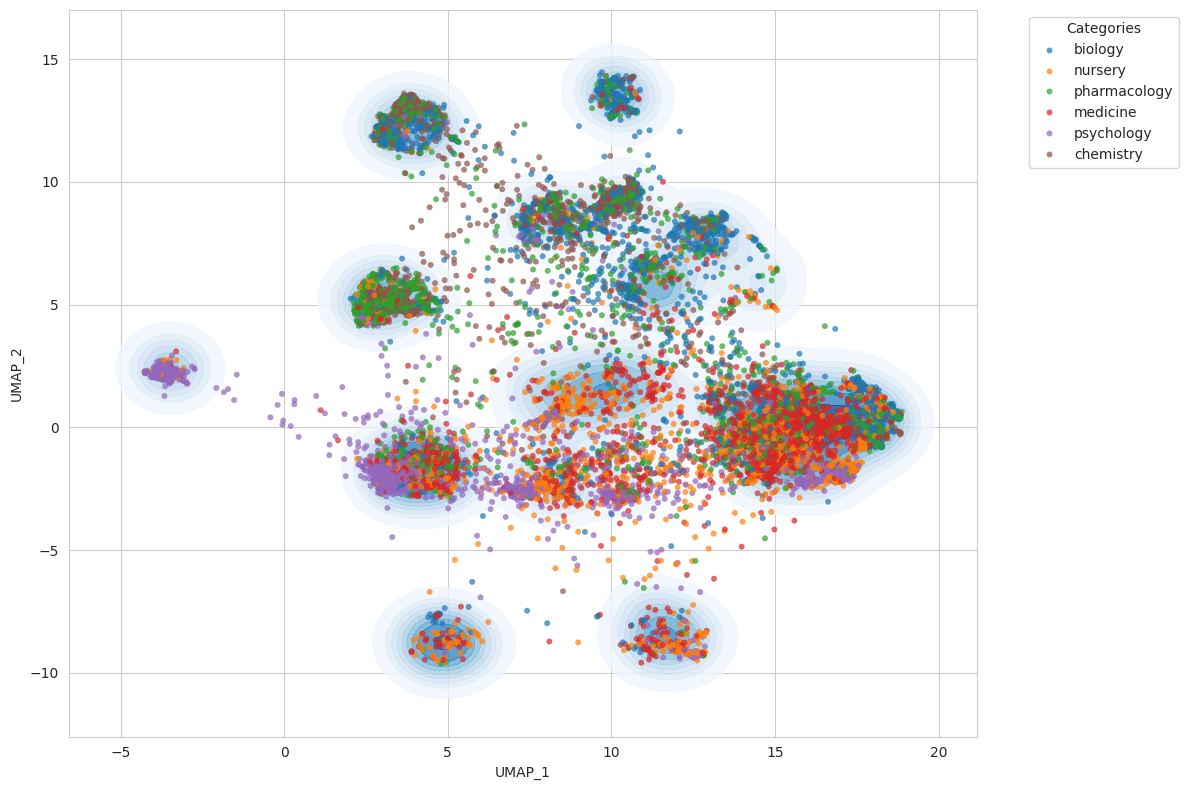}
        \caption{Scatter plot of HEAD-QA v2 questions by discipline overlaid on the corpus density map.}
        \label{subfig:books_and_queries}
    \end{minipage}
\end{figure*}

\paragraph{Assessing corpus alignment}
To evaluate the suitability of this corpus for our benchmark, Figures~\ref{subfig:textbooks}, \ref{subfig:texrbooks_global}, and \ref{subfig:books_and_queries} show a two-dimensional UMAP~\cite{mcinnes2020umapuniformmanifoldapproximation} projection of all 126k corpus fragments and the 12k HEAD-QA v2 questions.  
Distinct clusters correspond to individual textbooks, with minimal overlap.  
Importantly, most HEAD-QA v2 questions project into high-density corpus regions, indicating strong topical alignment.  
For example, psychology questions cluster around Psychiatry\_DSM-5 and Neurology\_Adams, while biology and pharmacology items align with related sources. These observations suggest that the corpus and retrieval setup may supply relevant contextual evidence, motivating their inclusion as a baseline for our benchmark.

\subsubsection{Selection via log-probabilities}
\label{subsec:logprob_classifier}

Unlike the previous methods, which require autoregressive text generation, this approach directly compares the probabilities that a language model assigns to each candidate answer sequence.  

Formally, let \(C = (c_1, c_2, \dots, c_n)\) represent the token sequence of a question and \(A_i = (a_1, a_2, \dots, a_m)\) the sequence corresponding to the \(i\)-th answer option.  
For each token \(a_j\), the model computes a conditional probability \(q_j = P(X_{n+j} \mid X_1=c_1, \dots, X_n=c_n, X_{n+1}=a_1, \dots, X_{n+j-1}=a_{j-1})\).  
The overall likelihood of an answer sequence is then defined as the geometric mean of its token probabilities,  
\(P(A_i) = \big(\prod_{j=1}^{m} q_j(a_j)\big)^{1/m}\).  
The model selects as correct the option that maximizes this probability, i.e., \(
= \arg\max_i P(A_i)\).  
Because multiplying many small probabilities can lead to numerical instability, all computations are performed in 32-bit precision.  
In addition, probabilities are evaluated in log-space to improve stability and efficiency, using the equivalent formulation  
\(\log P(A_i) = \tfrac{1}{m} \sum_{j=1}^{m} \log q_j(a_j)\).

\section{Experimental setup}

Performance is evaluated using three metrics:(1) accuracy, the proportion of correct answers; (2) the normalized exam score, based on the official Spanish medical exam scheme (three wrong answers cancel one correct) and normalized by total items; and (3) the unanswered ratio, the fraction of questions with no valid response.

\subsection{`Prompting Strategy' Evaluation}

\begin{figure}[phtb]
    \centering
    \includegraphics[width=0.4\textwidth]{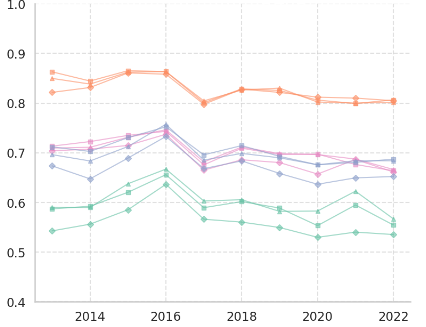}
\caption{Performance evolution over time for each model on the English subset under the prompting setup. Colors indicate model families: Mistral-7B (green), Mixtral-8x7B (blue), Llama-3.1-70B (orange), and Llama-3.1-8B (pink). Markers denote prompting strategies: squares for zero-shot, triangles for few-shot, and diamonds for CoT.}
    \label{fig:year_evolution_prompt}
\end{figure}

Table~\ref{tab:prompt_experiments_summary_tab} reports performance metrics for all prompting configurations (zero-shot, few-shot, and CoT).

\begin{table}[hbtp]
\centering
\scriptsize
\setlength{\tabcolsep}{2.5pt}
\begin{tabular}{|l|l|ccc|ccc|}
\hline
\multirow{2}{*}{\textbf{Prompt}} & \multirow{2}{*}{\textbf{Model}} & \multicolumn{3}{c|}{\textbf{English (en)}} & \multicolumn{3}{c|}{\textbf{Spanish (es)}} \\\cline{3-8}
 &  & Acc & Score & $P_\text{na}$ & Acc & Score & $P_\text{na}$ \\\hline
\multirow{4}{*}{Zero-shot} 
& Mixtral-8x7B & 70.59 & 66.97 & 2.03 & 66.01 & 60.43 & 4.94\\
& Mistral-7B & 59.55 & 52.61 & 4.82 & 52.79 & 43.56 & 3.25\\
& Llama-3.1-8B & 70.43 & 67.86 & 0.39 & 61.93 & 56.61 & 0.38\\
& Llama-3.1-70B & \textbf{83.15} & 84.16 & 0.43 & \textbf{83.27} & 84.14 & 0.40\\\hline

\multirow{4}{*}{Few-shot} 
& Mixtral-8x7B & 69.78 & 66.23 & 4.64 & 66.85 & 62.05 & 3.83\\
& Mistral-7B & 60.63 & 54.59 & 4.42 & 54.06 & 45.90 & 3.02\\
& Llama-3.1-8B & 70.49 & 68.24 & \textbf{0.36} & 62.58 & 57.44 & \textbf{0.24}\\
& Llama-3.1-70B & 82.90 & 84.14 & 0.51 & 83.24 & \textbf{84.41} & 0.38\\\hline

\multirow{4}{*}{CoT} 
& Mixtral-8x7B & 67.08 & 62.53 & 6.71 & 64.27 & 58.66 & 7.22\\
& Mistral-7B & 56.19 & 47.89 & 8.54 & 48.07 & 36.55 & 10.55\\
& Llama-3.1-8B & 69.13 & 66.50 & 5.55 & 61.05 & 55.30 & 6.11\\
& Llama-3.1-70B & 82.54 & \textbf{84.20} & 2.11 & 82.10 & 83.21 & 4.07\\\hline
\end{tabular}
\caption{Performance metrics (accuracy, exam score, and proportion of unanswered questions) for all prompting configurations (zero-shot, few-shot, and CoT) in English and Spanish. Best values per column are highlighted in bold.}
\label{tab:prompt_experiments_summary_tab}
\end{table}

Overall, performance is consistently higher in English than in Spanish across all configurations, except for Llama-3.1-70B, where results are equivalent. This confirms that models handle English—either natively or through translation—more effectively. The gap is particularly pronounced in smaller models, suggesting that limited capacity (at least to represent specialized healthcare knowledge) amplifies cross-lingual variability. In contrast, larger models show stronger generalization, narrowing the difference between languages. Model scale has a clear impact: accuracy and exam scores increase steadily with model size, while the proportion of unanswered questions decreases.

Regarding prompting strategies, zero-shot and few-shot approaches achieve comparable results, suggesting that providing a single example offers limited additional benefit given the models’ instruction tuning. Exploring the impact of example selection could be an interesting direction for future work. In contrast—perhaps unexpectedly—CoT prompting consistently reduces accuracy and increases non-response rates except for the Llama-3.1-70B, indicating that explicit reasoning steps may actually reduce performance in this healthcare domain.

Figure~\ref{fig:year_evolution_prompt} shows that English performance remains stable across exam years, with larger models outperforming smaller ones. English results are slightly higher than Spanish (not shown for space), and simpler prompting strategies get the most reliable outcomes.

\subsection{`RAG Strategy' Evaluation}

\begin{table}[hbtp]
\centering
\scriptsize
\setlength{\tabcolsep}{2.5pt}
\begin{tabular}{|l|l|ccc|ccc|}
\hline
\multirow{2}{*}{\textbf{Prompt}} & \multirow{2}{*}{\textbf{Model}} & \multicolumn{3}{c|}{\textbf{English (en)}} & \multicolumn{3}{c|}{\textbf{Spanish (es)}} \\\cline{3-8}
 &  & Acc & Score & $P_\text{na}$ & Acc & Score & $P_\text{na}$ \\\hline
\multirow{4}{*}{RAG} 
& Mixtral-8x7B & 69.80 & 66.25 & 4.67 & 66.90 & 62.07 & 3.91\\
& Mistral-7B & 56.63 & 49.11 & 2.14 & 49.61 & 39.70 & 2.30\\
& Llama-3.1-8B & 66.45 & 62.83 & 0.69 & 59.13 & 52.86 & 0.57\\
& Llama-3.1-70B & \textbf{82.45} & \textbf{83.13} & \textbf{0.32} & \textbf{82.52} & \textbf{83.03} & \textbf{0.22}\\\hline
\end{tabular}
\caption{Performance metrics (accuracy, exam score, and proportion of unanswered questions) for all RAG-based configurations in English and Spanish.}
\label{tab:rag_experiment_results}
\end{table}

Table~\ref{tab:rag_experiment_results} presents the performance metrics for the models that used RAG to condition their prompt.

Overall, results show that incorporating retrieved context through RAG does not lead to consistent improvements over standard prompting. Performance remains slightly higher in English than in Spanish. Larger models benefit the most from RAG, maintaining competitive accuracy and lower non-response rates, while smaller models tend to degrade when exposed to noisy or weakly relevant evidence.

\begin{figure}[hpbt]
    \centering
    \includegraphics[width=0.45\textwidth]{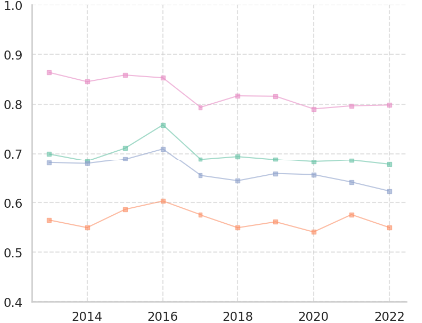}
\caption{Performance evolution over time for each model in English under the RAG setup. Colors indicate model families: Mixtral-8x7B (green), Mistral-7B (orange), Llama-3.1-8B (blue), and Llama-3.1-70B (pink).}
    \label{fig:year_evolution_rag}
\end{figure}

Compared to the zero-shot baseline, RAG gets slightly lower scores in both languages. This suggests that the retrieved passages are not always effectively integrated into the generation process, that the model can often rely on its internal knowledge instead, or that the retrieved information is not sufficiently relevant. The weak correlation between retrieval relevance (see \S \ref{subsec:rag_proposal}) and answer correctness ($r = 0.07$) further supports this interpretation: model performance appears to depend primarily on internal knowledge rather than external evidence. Yearly trends remain stable across models and languages, closely mirroring those observed in the prompting experiments, as shown in Figure \ref{fig:year_evolution_rag}.

\subsection{`Log-probability' Evaluation}

Table~\ref{tab:logprob_classifier_experiment_results} reports accuracy and normalized exam scores for this setup. By design, the unanswered ratio is $0\%$, as models are required to select one option per question. Despite this, scores drop notably compared to prompting-based approaches, indicating that direct likelihood evaluation is less effective for multiple-choice reasoning. Performance remains consistently higher in English than in Spanish, with the gap being more pronounced in smaller models. Larger models mitigate this difference, maintaining more stable accuracy across languages. The observed performance gap can also be attributed to the independent evaluation of each option: without jointly considering all alternatives, models lose the elimination-based reasoning that benefits prompting approaches.

\begin{figure}[hpbt]
    \centering
    \includegraphics[width=0.45\textwidth]{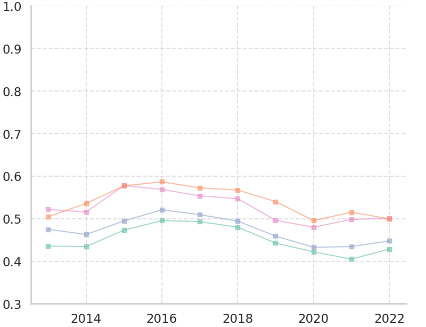}
\caption{Performance evolution over time for each model in English under the probability-based selection setup. Colors indicate model families: Llama-3.1-8B (green), Llama-3.1-70B (orange), Mistral-7B (blue), and Mixtral-8x7B (pink).}
    \label{fig:year_evolution_prob}
\end{figure}

As shown in Figure~\ref{fig:year_evolution_prob}, yearly trends remain stable, with only minor fluctuations after 2018. Although this method minimizes resource usage—since no text generation is involved—the efficiency gain does not compensate for the accuracy loss, limiting its practical value for HEAD-QA v2.

\begin{table}[hbtp]
\centering
\scriptsize
\setlength{\tabcolsep}{5pt}
\begin{tabular}{|l|l|cc|cc|}
\hline
\multirow{2}{*}{\textbf{Strategy}} & \multirow{2}{*}{\textbf{Model}} & \multicolumn{2}{c|}{\textbf{English (en)}} & \multicolumn{2}{c|}{\textbf{Spanish (es)}} \\\cline{3-6}
 &  & Acc & Score & Acc & Score \\\hline
\multirow{4}{*}{Prob-based} 
& Mixtral-8x7B & 52.84 & 44.06 & 47.87 & 37.60\\
& Mistral-7B & 47.86 & 37.49 & 39.42 & 27.69\\
& Llama-3.1-8B & 45.25 & 34.31 & 37.82 & 24.80\\
& Llama-3.1-70B & \textbf{54.15} & \textbf{46.04} & \textbf{51.42} & \textbf{42.12}\\\hline
\end{tabular}
\caption{Performance metrics (accuracy and exam score) for the probability-based selection strategy (Section~\ref{subsec:logprob_classifier}) in English and Spanish.}
\label{tab:logprob_classifier_experiment_results}
\end{table}

\section{Discussion}
\label{sec:discussion}

The results reveal consistent trends across model families, highlighting how architectural scale, language, and methodological design shape performance in HEAD-QA~v2. Model size emerges as the most decisive factor: Llama-3.1-70B consistently achieves the highest accuracy and normalized exam scores, while smallest models performs lowest across all metrics. These results align with broader findings in LLM evaluation, where scaling enhances both factual recall and reasoning stability.

Language effects are present but moderate, with smaller models showing slightly reduced performance in Spanish. This may stem from differences in tokenization efficiency, knowlegde integration, and from weaker multilingual representations in smaller architectures, which maybe be less robust to lexical and syntactic variability across languages.

Methodologically, neither more elaborate prompting (few-shot or CoT) nor retrieval-augmented generation produces consistent improvements. In some cases, these strategies even reduce performance, suggesting that additional contextual input can introduce noise or divert the model from leveraging its internal knowledge. Considering their higher computational and developmental costs, such methods offer limited benefit in this setting.

Finally, the probability-based answer selection strategy performs notably worse than generation-based approaches. Since each option is scored independently, the model cannot perform the comparative reasoning and contextual alignment typical of human multiple-choice problem-solving, resulting in systematic accuracy drops.

\section{Conclusion}

This work introduced HEAD-QA v2, a new large-scale, multilingual benchmark designed to evaluate complex reasoning in the biomedical domain. Through extensive experiments across multiple modern large language models and inference strategies, we established empirical baselines and analyzed the factors that most influence performance. Our findings indicate that, for highly specialized biomedical question answering, the intrinsic knowledge and reasoning capacity of the language model play a far greater role than the sophistication of the inference strategy. Techniques such as RAG and CoT prompting, while successful in other domains, did not obtain consistent gains in this setting and introduced additional computational and implementation overhead. Overall, improvements on HEAD-QA v2 seem more closely tied to scaling and refining the underlying models than to increasing inference complexity, though alternative strategies may still offer potential for future exploration.

\section*{Limitations}
\label{sec:limitations}

This study did not include evaluations with frontier proprietary LLMs such as GPT-4, Claude, or Gemini, primarily due to funding resources to access APIs. Consequently, the results reflect trends among open-access models up to 70B parameters.

Additionally, while the English translations were automatically generated and reviewed for terminological consistency, large-scale human validation was not feasible. Minor translation inconsistencies could therefore influence model performance, especially in domain-specific terminology.

Another limitation concerns the scope of the benchmark itself. HEAD-QA~v2 focuses on multiple-choice biomedical questions, which represent only a subset of complex reasoning skills.

\section*{Ethical Considerations}
\label{sec:ethics}

HEAD-QA~v2 is based on publicly available examination questions designed for healthcare education, containing no personal or patient data. Nevertheless, the dataset and experiments involve content related to medical knowledge, and outputs from large language models should not be interpreted as clinical advice.

All experiments were conducted with open-access models and publicly available data, ensuring reproducibility and compliance with data use terms. We acknowledge that automatic translation and model-generated text may propagate biases or inaccuracies, and encourage caution when using the dataset or models in real-world or educational healthcare contexts.

\section*{Acknowledgments}

We acknowledge grants GAP (PID2022-139308OA-I00) funded by MICIU/AEI/10.13039/501100011033/ and ERDF, EU; LATCHING (PID2023-147129OB-C21) funded by MICIU/AEI/10.13039/501100011033 and ERDF, EU; and TSI-100925-2023-1 funded by Ministry for Digital Transformation and Civil Service and ``NextGenerationEU'' PRTR; as well as funding by Xunta de Galicia (ED431C 2024/02), and 
CITIC, as a center accredited for excellence within the Galician University System and a member of the CIGUS Network, receives subsidies from the Department of Education, Science, Universities, and Vocational Training of the Xunta de Galicia.
Additionally, it is co-financed by the EU through the FEDER Galicia 2021-27 operational program (Ref. ED431G 2023/01). This research project was made possible through the access granted by the Galician Supercomputing Center (CESGA) to its supercomputing infrastructure. The supercomputer FinisTerrae III and its permanent data storage system have been funded by the NextGeneration EU 2021 Recovery, Transformation and Resilience Plan, ICT2021-006904, and also from the Pluriregional Operational Programme of Spain 2014-2020 of the European Regional Development Fund (ERDF), ICTS-2019-02-CESGA-3, and from the State Programme for the Promotion of Scientific and Technical Research of Excellence of the State Plan for Scientific and Technical Research and Innovation 2013-2016 State subprogramme for scientific and technical infrastructures and equipment of ERDF, CESG15-DE-3114. Finally, we are grateful to the funding of \emph{Consellería de Educación, Ciencia, Universidades e Formación Profesional (Xunta de Galicia - Convenio para o desenvolvemento de accións estratéxicas de I+D+i 2025-2026).}

\section{Bibliographical References}\label{sec:reference}

\bibliographystyle{lrec2026-natbib}

\bibliography{lrec2026-example}


\end{document}